\documentclass[11pt]{article}

\usepackage[utf8]{inputenc}
\usepackage[T1]{fontenc}
\usepackage{lmodern}
\usepackage{microtype}
\usepackage{amsmath,amsfonts}
\usepackage{graphicx}
\usepackage{booktabs}
\usepackage{hyperref}
\usepackage{enumitem}
\usepackage{geometry}
\usepackage{parskip}

\title{\textbf{Auditable Decision Models with Learned Abstention and Real-Time Steering}}

\author{
Sankaranarayanan Palamadai Chandrasekaran \\
Simple Machine Mind \\
\texttt{sankar@smsquared.ai}
}

\date{}

\begin{document}

\maketitle

\begin{abstract}
Production AI systems often operate with incomplete, conflicting, or insufficient evidence. Forced classifiers collapse such cases into action labels, while generative systems can produce outputs that are difficult to interpret as auditable execution decisions. We study operational decision control for AI systems, where uncertainty must be explicitly routable, policy-governed, and auditable rather than hidden inside forced predictions or free-form generation. We present EvaluatorDPT, a bounded decision-control model that predicts YES, NO, or TBD, where TBD is learned as a deferral outcome rather than added only as a post-hoc confidence rule. The model uses a transformer encoder with a primary bounded-decision head and structured auxiliary channels for values and emotions/sentiments. The interface is domain-agnostic in form: a deployment domain supplies evidence and policy thresholds, while the model emits a bounded distribution that can be controlled at inference time through recorded operating thresholds and, when validated, auxiliary semantic signals. For the evaluated model version, we report decision performance on held-out validation and test splits; auxiliary emotion metrics are omitted because the emotion head is disabled for this evaluation. On the held-out test split (n=44,597), the model achieves Accuracy = 0.8260 and Macro F1 = 0.8252, with per-class F1 of 0.8314 (YES), 0.8486 (NO), and 0.7956 (TBD). The evaluation record also includes calibration evidence (ECE = 0.0338 on validation), threshold-sweep outputs, multi-seed stability checks, confusion matrices, and reproducibility commands. Our main contribution is a bounded execution interface in which deferral is learned, inference-time routing remains inspectable, auxiliary signals provide a path to auditable behavior control, and evaluation evidence supports external review.
\end{abstract}

\section{Introduction}

Many deployed systems now connect language inputs to downstream actions. In such settings, classification accuracy alone is not the central question; the system must also decide whether to act, block, or defer. Inputs may be incomplete, contradictory, or ambiguous. A model that must always choose YES or NO can conceal uncertainty in cases where an operator would prefer review.

The issue appears when language models or automated agents are connected to business workflows, policy enforcement, moderation, approvals, safety checks, or other action-taking systems. In these environments, a model output is not merely a label. It becomes a control signal. A false YES may trigger an unsafe or unauthorized action; a false NO may block a valid action; and a failure to defer may hide uncertainty from the surrounding workflow. A useful decision model therefore needs an explicit representation of uncertainty, rather than a confidence score attached to a forced label.

The model is designed for this setting. It is not a general-purpose language generator or a replacement for domain-specific review. It is a bounded decision interface that converts a context signal into a constrained routing decision with confidence and supporting evaluation evidence. Downstream policies can inspect the decision distribution, choose operating thresholds, route deferred cases, and preserve a review trail.

The proposed formulation models decision control as a bounded three-way output space:
\[
\mathcal{Y}_d = \{\text{YES}, \text{NO}, \text{TBD}\}.
\]
The TBD outcome is not merely a post-hoc rejection threshold. It is a learned class that allows the model to internalize deferral boundaries during training. Thresholding remains useful at deployment time, but it is applied on top of a model that already represents ambiguity as part of the label space.

The contribution of this work is \textbf{learned deferral as an auditable operational decision-control formulation}. The model does not generate free-form decisions. It emits bounded labels, confidence scores, and structured auxiliary signals, while the evaluation package provides the evidence needed to inspect behavior: calibration, threshold sweeps, multi-seed stability, confusion matrices, and reproducibility materials. The same interface can be applied across domains because the output is a policy-facing decision interface rather than a domain-specific action generator; however, each new domain still requires its own calibration and validation. This supports inference-time decision control: an operator can change threshold policy, deferral routing, or auxiliary-signal use without modifying the model weights, provided those changes are recorded as operating policy.

This framing is intentionally narrower than many general language-model evaluations. The paper does not argue that a single model can resolve all downstream governance questions. Instead, it studies a smaller but operationally important problem: how to expose an explicit, learned deferral decision in a form that can be evaluated, thresholded, logged, and reviewed. The design goal is not maximal automation. It is controlled routing under ambiguity.

\paragraph{Contributions.}
\begin{itemize}[leftmargin=1.5em]
  \item We formulate decision routing as bounded control over YES, NO, and learned TBD deferral.
  \item We show how operating thresholds turn the model distribution into an auditable decision-routing interface.
  \item We present a transformer-based multi-head architecture that separates primary decision prediction from structured auxiliary context signals.
  \item We define auxiliary value and emotion/sentiment channels as decision-relative semantic signals that can support auditable policy control when they are enabled and separately validated.
  \item We describe targeted boundary refinement for hard ambiguity cases, including negation, contradiction, lexical-overlap traps, and modal/hedged statements.
  \item We evaluate the reported model version on full validation and test splits and report per-class behavior, calibration, threshold-sweep evidence, and stability checks.
  \item We organize evaluation results, calibration data, threshold sweeps, and reproducibility materials into a coherent review structure.
\end{itemize}

\section{Problem Formulation}

This work studies a decision-control setting rather than a standard text-classification setting. A model receives a context signal $x$ and must output a decision that can be consumed by a downstream workflow. The decision may authorize an action, block an action, or defer the action for additional review. The output is therefore not only a semantic label; it is an interface between language understanding and operational policy.

We define the bounded decision space as
\[
\mathcal{Y}_d = \{\text{YES}, \text{NO}, \text{TBD}\}.
\]
YES denotes sufficient support for action, NO denotes sufficient support for rejection or blocking, and TBD denotes insufficient certainty for either direct action or direct rejection. In deployment, TBD may route to human review, a policy engine, evidence collection, or another conservative fallback.

This formulation differs from binary classification with a reject threshold in two ways. First, deferral is represented in the supervised label space, so ambiguous examples can influence the learned representation. Second, thresholding remains available as an operating-policy layer over a three-way decision distribution. The model can learn the semantic structure of deferral, while deployment owners can still select thresholds that reflect local risk tolerance.

The practical requirement is not only high aggregate performance. A decision-control model should also expose the evidence needed to review its behavior: per-class outcomes, confusion patterns, calibration, threshold tradeoffs, stability under repeated evaluation, and model identity. These requirements motivate the audit-evidence reporting structure described later in the paper.

\subsection{Operational Risk}

The operational risk of a bounded decision model depends on both the predicted label and the action policy attached to that label. Let $a(g(x))$ denote the downstream action selected after applying a routing rule $g$. A simple deployment loss can be written as
\[
R(g) = \mathbb{E}_{(x,y)} \left[ C(y, g(x)) \right],
\]
where $C$ assigns different costs to different errors. In a decision-routing setting, confusing a true TBD case with YES or NO can be more costly than routing a true YES or NO case to TBD, because over-commitment may trigger an unsupported action while false deferral usually triggers review. This asymmetry motivates reporting false-action behavior, high-confidence errors, and threshold sweeps in addition to Macro F1.

This paper does not estimate application-specific costs. Instead, it reports classwise errors and coverage-sensitive operating points so that downstream users can choose thresholds consistent with their own cost matrix. This keeps the model evaluation separate from deployment-specific policy selection.

\subsection{Learned Deferral and Reject-Threshold Abstention}

Let $p_\theta(y \mid x)$ be the model distribution over $\mathcal{Y}_d$. In a standard reject-threshold design, a binary classifier first estimates $p_\theta(y \in \{\text{YES}, \text{NO}\}\mid x)$ and then abstains when confidence, entropy, or margin fails a threshold. Deferral is therefore an operating rule applied after the classifier has learned a binary decision boundary.

In this work, deferral is part of the supervised decision space. The training objective includes examples labeled TBD, so the model can assign probability mass directly to an insufficient-evidence outcome:
\[
\hat{y} = \arg\max_{y \in \{\text{YES},\text{NO},\text{TBD}\}} p_\theta(y \mid x).
\]
A deployment threshold $\tau_y$ may still be applied:
\[
g(x) =
\begin{cases}
\arg\max_y p_\theta(y \mid x), & \exists y: p_\theta(y \mid x) \ge \tau_y,\\
\text{TBD}, & \text{otherwise.}
\end{cases}
\]
The distinction is not that thresholds are avoided. The distinction is that thresholds operate over a distribution that already contains a learned deferral class. The experiments below therefore evaluate two separate questions: the value of a learned three-way decision space at full coverage, and the operating tradeoff obtained by applying reject thresholds to model confidence.

\section{What Is Novel in This Work}

The novelty is not the use of a transformer encoder by itself. The contribution lies in the way the model and evaluation protocol are organized around bounded decision control.

\paragraph{Learned deferral.}
TBD is treated as a supervised outcome, not only as a threshold applied after a YES/NO prediction. This allows ambiguous and insufficient-evidence examples to participate in training. The model can therefore learn patterns that correspond to deferral rather than treating all low-confidence regions as equivalent.

\paragraph{Separation of decision and operating policy.}
The model emits a bounded decision distribution, while deployment thresholds remain external and inspectable. This separates the representation of uncertainty from the choice of how conservatively the system should act in a particular environment.

\paragraph{Inference-time decision control.}
The bounded distribution is also an inference-time control interface. A deployment can increase or decrease automation by changing class thresholds, route low-margin predictions to TBD, or make TBD the default fallback under higher-risk conditions. These changes affect routing behavior at inference time without retraining or replacing the model. The routing policy is auditable because it can be recorded as thresholds, fallback rules, and enabled auxiliary channels rather than embedded in free-form generation.

\paragraph{Domain-agnostic control interface.}
The output space is intentionally independent of any single downstream domain. A moderation system, approval workflow, compliance review, or financial triage system can attach different operating thresholds and escalation rules to the same bounded labels. This does not imply zero-shot production readiness across domains; it means that the interface separates domain policy from the model's learned decision distribution, making domain transfer a validation question rather than a redesign of the output space.

\paragraph{Auxiliary policy control.}
The multi-head design allows auxiliary value and emotion/sentiment signals to be treated as structured decision-relative semantic channels. When those channels are enabled and validated, they can be logged, thresholded, or used by a downstream policy engine to control routing behavior in real time via versioned thresholds and routing policy, without changing the model weights. In the evaluated model version, empirical claims remain limited to the decision head because the emotion head is disabled for this evaluation; auxiliary policy control is therefore presented as an architectural capability and a future empirical claim, not as a validated result for the reported model.

\paragraph{Release-level auditability.}
The evaluation package includes calibration data, threshold-sweep outputs, multi-seed stability checks, confusion matrices, reproducibility commands, and model/data pointers. These materials make the evaluated behavior reviewable. This is a different standard from publishing only model weights and a top-line score.

\paragraph{Boundary-focused refinement.}
The development process identifies deferral boundaries as the key failure region and uses targeted boundary examples rather than undifferentiated data growth. This is especially relevant for negation, contradiction, lexical-overlap traps, long-text stress cases, and hedged statements.

Together, these choices define the scope of the paper. The model is presented as a bounded execution interface for decision routing, not as a new language-model backbone. The empirical question is whether a learned deferral class, evaluated with classwise and calibration evidence, can support controlled decision workflows.

\subsection{Design Principles}

The work is guided by six design principles for auditable decision routing:
\begin{enumerate}[leftmargin=1.5em]
  \item Uncertainty should be explicit, not hidden inside a forced YES/NO decision.
  \item Deferral should be represented as a first-class outcome when training data supports it.
  \item Operating thresholds should remain external to the model weights and be recorded as policy.
  \item Confidence should be accompanied by calibration evidence.
  \item Evaluation should report classwise behavior, not only aggregate accuracy.
  \item The evaluated model version, split, threshold policy, and commands should be reconstructable.
\end{enumerate}

\subsection{Auditability Properties}

The auditability claim in this paper refers to review properties of the evaluated decision system, not to a guarantee of correctness. The relevant properties are:
\begin{itemize}[leftmargin=1.5em]
  \item \textbf{Traceability}: the evaluated model version, dataset split, metrics, and threshold policy can be identified.
  \item \textbf{Replayability}: the evaluation can be rerun from recorded commands, logits, labels, and configuration.
  \item \textbf{Threshold reconstructability}: the final routing rule can be reconstructed from stored class thresholds and fallback behavior.
  \item \textbf{Calibration observability}: confidence behavior is reported through reliability bins and ECE rather than assumed from softmax scores.
  \item \textbf{Policy separability}: the model distribution and deployment threshold policy are documented separately.
\end{itemize}

These properties define a practical review standard for decision-routing systems. They do not replace external validation, domain-specific governance, or prospective monitoring.

\section{Related Work}

The proposed model is related to selective classification and reject-option models, which abstain from prediction under uncertainty \cite{elyaniv2010selective, geifman2017selective, franc2023reject}. In many such systems, abstention is applied after prediction using confidence thresholds. Here, the model instead learns a deferral label directly while still preserving threshold sweeps for operating-point selection. A learned TBD class allows ambiguous examples to shape the representation during training, while thresholding remains a deployment policy rather than the only source of abstention.

The architecture builds on transformer language models such as BERT \cite{devlin2019bert} and multi-task learning \cite{caruana1997multitask}. Here, auxiliary outputs are treated as structured context signals available at inference time rather than only as training regularizers.

The auxiliary channels also differ from conventional sentiment or lexical classification tasks. In this work, value and emotion/sentiment signals are interpreted relative to the decision event. They are intended to encode decision-relevant semantic context rather than surface affect alone. This distinction is important because a sentence can contain positive or negative language without providing sufficient support for a YES or NO decision.

Neural confidence scores can be miscalibrated \cite{guo2017calibration}, so the evaluation pairs decision outputs with calibration evidence and threshold-sweep outputs. This is relevant for governance because the operating policy should be inspectable rather than implicit.

The paper is also related to model cards and governance-oriented reporting practices. Such documents often record intended use, limitations, and evaluation results after training. In this work, these materials are treated as part of the evaluation record: the model is accompanied by structured metrics, calibration data, threshold sweeps, stability checks, and reproducibility commands.

\section{System Overview}

The model receives a context signal and returns:
\begin{itemize}[leftmargin=1.5em]
  \item a bounded decision: YES, NO, or TBD;
  \item decision confidence derived from the primary decision head;
  \item structured auxiliary signals for values and emotions/sentiments, where enabled and validated;
  \item evaluation evidence that supports threshold selection and audit.
\end{itemize}

The intended deployment pattern is conservative. YES and NO represent bounded action signals. TBD represents a case for human review, a policy engine, additional evidence collection, or another fallback. Because the outputs are bounded, downstream systems can log the decision distribution and threshold policy without parsing generated text.

The system boundary is therefore clear. The model does not execute the downstream action, resolve policy conflicts, or replace review. It provides a decision-control signal that can be consumed by another system. A deployment can use the raw argmax decision, enforce a minimum confidence margin, route all TBD cases to review, or apply a more conservative threshold policy derived from the threshold-sweep outputs. This keeps the model output separate from the operational decision rule.

\subsection{Operational Interface}

In an operational setting, each prediction produces a record containing the input identifier, model version, model identifier, decision probabilities, selected operating threshold, final routed label, and any auxiliary channels enabled for the model version. This structure is important because reviewability depends on reconstructing not only the model score but also the policy applied at the time of decision.

The interface also supports conservative fallback and inference-time decision control. A downstream system can treat low-confidence YES or NO predictions as TBD without retraining the model. Conversely, if a use case tolerates more automation, it can lower the deferral threshold while preserving the same trained model. The same mechanism can be changed by operating condition: for example, a high-risk workflow can increase TBD routing during audit periods, while a lower-risk workflow can accept more direct YES or NO decisions. Accordingly, threshold selection is treated as part of the recorded evaluation evidence rather than as an implementation detail.

Auxiliary channels extend this control pattern when they are validated for a model version. A policy engine can use auxiliary value or emotion/sentiment scores as additional routing conditions, for example by requiring review when the primary decision is high-confidence but an auxiliary signal indicates a sensitive or adversarial context. This paper does not claim validated auxiliary policy control for this model version; it defines the interface and reports the validated decision-head evidence.

\subsection{Operational Lifecycle}

The intended lifecycle separates model estimation from policy: thresholding and routing policy remain external to the model weights. First, the model emits a bounded decision distribution and any validated auxiliary channels. Second, a policy layer applies versioned thresholds and fallback rules. Third, the routing layer returns YES, NO, or TBD to the downstream workflow. Fourth, the audit record stores the model identifier, policy version, class probabilities, final routed label, and enabled auxiliary channels. Finally, governance review can replay decisions, compare policy versions, and change operating thresholds without retraining the model. This lifecycle is the operational object of study: the model is an implementation component inside a governed routing architecture.

\begin{figure}[htbp]
\centering
\includegraphics[width=0.9\linewidth]{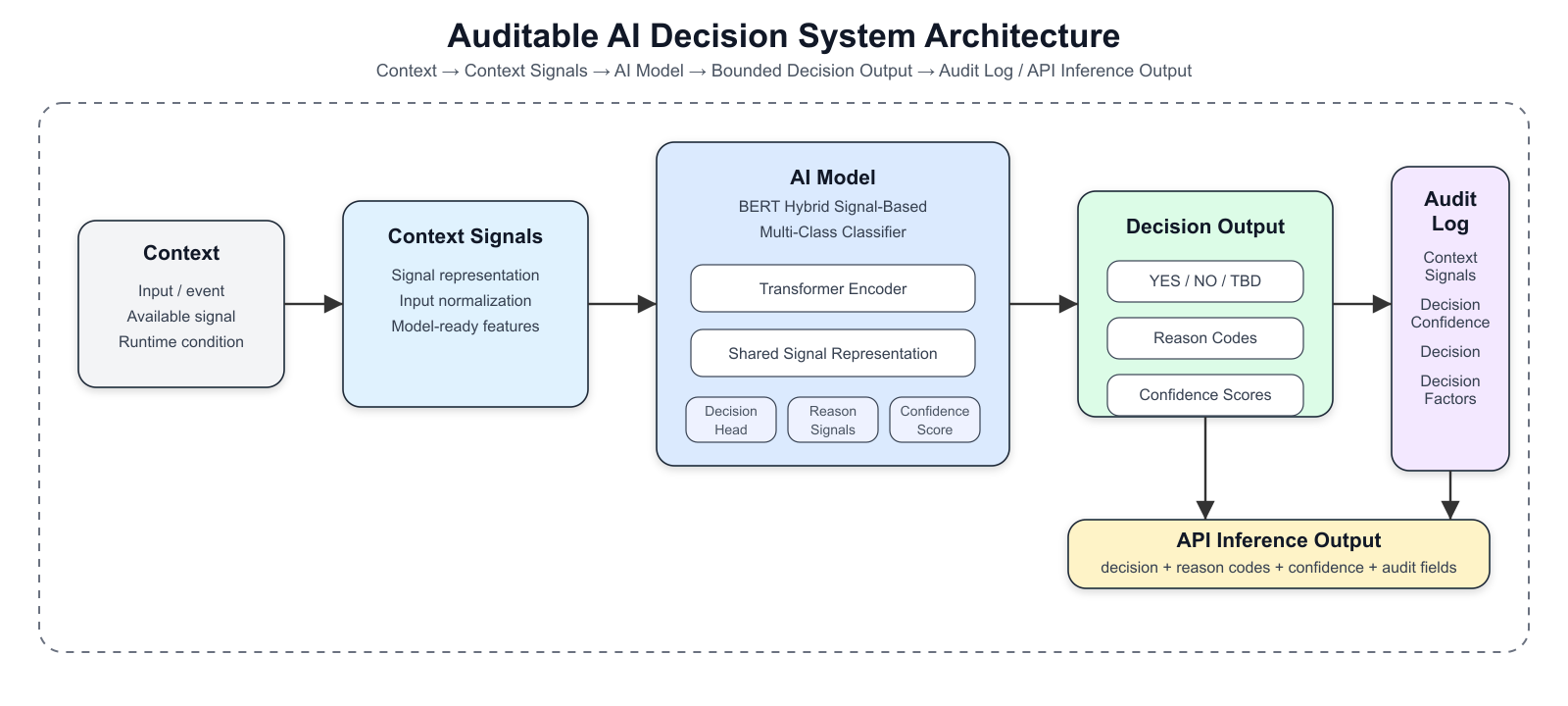}
\caption{System architecture: tokenized decision event to shared transformer encoder, bounded decision output, confidence, and structured auxiliary channels.}
\end{figure}

\section{Methodology}

\subsection{Bounded Decision Formulation}

Let $x$ be a decision-event input. The model predicts a primary decision distribution over $\mathcal{Y}_d = \{\text{YES}, \text{NO}, \text{TBD}\}$. The TBD label represents explicit deferral under insufficient semantic certainty. This differs from binary classification followed by confidence filtering because the model learns examples of deferral as part of the supervised task.

Deployment policies may still choose operating thresholds over the predicted probabilities. In this formulation, thresholding is an operating policy layered over a bounded decision model, not the sole mechanism by which uncertainty is represented.

This formulation gives the system two mechanisms for risk control. The first is learned: ambiguous or insufficient contexts can be labeled TBD during training. The second is operational: deployment owners can choose thresholds over the learned distribution to match risk tolerance. A high-assurance setting may route more cases to TBD, while a lower-risk setting may accept a smaller margin for YES or NO. The model separates the representation of uncertainty from the policy decision about how much uncertainty is acceptable. Inference-time control is therefore expressed as a change in recorded policy rather than as an untracked change to model behavior.

\subsection{Architecture}

The architecture uses a BERT-base encoder \cite{devlin2019bert} with a maximum sequence length of 128 tokens. The shared encoder feeds task-specific heads:
\begin{itemize}[leftmargin=1.5em]
  \item a primary three-class decision head;
  \item a 10-label value head;
  \item a 28-label emotion/sentiment head.
\end{itemize}

For the reported model version, the decision head is the validated output analyzed in this paper. The emotion head is disabled for this evaluation, so emotion metrics are intentionally not claimed. The auxiliary-head architecture remains part of the model design, but this paper scopes empirical claims to the validated decision behavior and evaluation evidence.

\subsection{Training Configuration}

The reported model starts from the preceding fixed-split decision model and applies one boundary-refinement epoch. The training configuration uses BERT-base-uncased, maximum sequence length 128, batch size 32, evaluation batch size 64, cosine scheduling with warmup ratio 0.05, label smoothing 0.05, and loss weights 1.0, 0.3, and 0.3 for the decision, value, and emotion/sentiment heads respectively. Encoder layers 9--11 are unfrozen with layer-wise learning rates $1.0{\times}10^{-6}$, $7.5{\times}10^{-7}$, and $5.0{\times}10^{-7}$. The global learning rate is $5.0{\times}10^{-6}$ and the recorded seed is 42. The emotion head is disabled during this evaluation, so no emotion-head metric is reported for the evaluated model.

\subsection{Data and Source Scope}

The evaluated corpus is a curated decision-event dataset built primarily from natural-language inference and stance-style material. The construction scripts map entailment, contradiction, and neutral or insufficient-evidence cases into the bounded YES/NO/TBD decision space. The sources identified in the construction and annotation scripts are SNLI \cite{bowman2015snli}, MultiNLI \cite{williams2018broad}, and SemEval-style rumor/stance material. The repository does not redistribute raw training data. Instead, it provides dataset documentation, processing notes, evaluation outputs, and reproducibility pointers sufficient for review of the evaluated model.

The dataset should be interpreted as a research and evaluation corpus for bounded decision behavior, not as a universal production benchmark. The paper evaluates learned deferral under the reported distribution; it does not claim universal generalization to all domains.

The dataset includes three broad kinds of material. First, natural-language inference examples expose entailment, contradiction, and neutral/insufficient-evidence patterns. Second, stance-style examples contribute support, denial, query, and uncertainty patterns that are mapped into the bounded decision space. Third, targeted boundary material is added for cases that were difficult in earlier diagnostics, including number changes, role changes, negation, high-overlap contradiction, low-overlap support, long-context stress, and modal or hedged language. These categories are described at a level suitable for publication; reuse of upstream datasets must follow the licenses of the original sources.

\begin{table}[htbp]
\centering
\begin{tabular}{lrrr}
\toprule
Split & YES & NO & TBD \\
\midrule
Validation & 14,962 & 15,534 & 13,908 \\
Test & 14,883 & 15,650 & 14,064 \\
\bottomrule
\end{tabular}
\caption{Decision-class support for the evaluated held-out splits, confirmed from the tokenized evaluation files used for scoring.}
\end{table}

The evaluated validation and test splits contain 44,404 and 44,597 examples respectively. These counts and class supports are taken from the tokenized evaluation files used for the reported scoring run. The training corpus contains approximately 419K examples after targeted hard-case construction. The boundary-refinement subset contains 323 training examples mined from prior diagnostic errors; it is used only for one sharpening epoch and is less than 0.1\% of the training volume.

\subsection{Annotation and Construction Transparency}

The decision labels are derived from a natural-language inference formulation. Entailment maps to YES, contradiction maps to NO, and neutral or insufficient evidence maps to TBD. No new project-specific inter-annotator agreement study was conducted for the reported model version. The project inherits the annotation assumptions of the source NLI datasets, including reported agreement for SNLI and MultiNLI, and applies documented correction passes where diagnostics identified systematic label errors.

The main dataset construction steps are also recorded. The initial cleaned corpus contained approximately 814K examples. A subsequent construction step reduced the corpus to the NLI-compatible subset, producing the largest quality gain in the dataset construction process. A canonical fixed-split baseline was then established. The final training split was expanded by adding targeted hard cases and capped pattern boosts while keeping the evaluated validation and test splits fixed. The final CSV files do not preserve a row-level source column, so the paper does not claim an exact per-source row breakdown. The exact per-source breakdown inside the 323-row boundary-refinement subset was likewise not preserved in the final CSV; this is recorded as a dataset-transparency limitation rather than inferred after the fact.

\subsection{Label Semantics}

The labels are defined by decision sufficiency rather than by surface sentiment. YES means the input provides enough support for the affirmative action or decision. NO means the input provides enough support to reject or block. TBD means the input lacks sufficient support for either direct action or direct rejection under the modeled policy. This distinction matters because a positive-sounding sentence may still be TBD if it does not provide enough evidence, and a negative-sounding sentence may be YES if it supports an affirmative control decision in context.

The auxiliary value and emotion/sentiment channels are designed to represent structured context, not to replace the primary decision label. Their intended semantics are decision-relative: the labels describe values and affective context as they bear on the decision event, not merely the linguistic tone of the text. The project documentation references established value, sentiment, emotion, and social-norm resources when describing auxiliary source scope \cite{schwartz2012overview,rosenthal2017semeval,malo2014good,barbieri2020tweeteval,demszky2020goemotions,ziems2023normbank,hoover2020mftc,poria2019meld,rashkin2019empathetic,sap2020socialbias,hendrycks2021ethics,go2009twitter,maas2011learning}. For example, lexical positivity does not imply a YES label, and lexical negativity does not imply a NO label. The relevant question is whether the text provides decision-sufficient support, rejection, or uncertainty.

For the evaluated model, the empirical claims are restricted to the decision head because the emotion head is disabled in this evaluation. This avoids overstating auxiliary performance and keeps the paper aligned with the validated evaluation evidence. Future auxiliary-head evaluation should therefore test decision-relative semantic alignment rather than only agreement with surface sentiment or keyword-based labels.

Representative label examples are:
\begin{itemize}[leftmargin=1.5em]
  \item YES: ``All employees received a 10\% pay raise this year'' supports ``Employee salaries increased this year.''
  \item NO: ``The company reported zero profit for the quarter'' contradicts ``The company was profitable this quarter.''
  \item TBD: ``The manager approved one new project this quarter'' does not establish ``The manager approved all new projects this quarter.''
  \item TBD: ``The policy may allow expedited review in limited cases'' does not establish ``The request is approved.''
\end{itemize}

\subsection{Boundary Refinement}

A recurring challenge in bounded decision systems is the boundary between action and deferral. The reported model uses the evaluated corpus with a targeted boundary-refinement subset focused on hard semantic regions such as contradiction, negation, low lexical overlap entailment, high lexical overlap contradiction, long-text stress cases, and hedge/modal ambiguity. The goal is to improve boundary behavior using targeted examples rather than relying only on broad dataset scale.

This design reflects an empirical lesson from the development process: additional data is most useful when it targets the failing boundary. The boundary-refinement subset focuses on cases where surface overlap can be misleading, where negation changes the decision, where numbers or roles alter entailment, or where a statement is too hedged to support a direct YES or NO. These cases matter for learned deferral because the model must distinguish confident action from insufficient evidence.

\section{Evaluation Setup}
\subsection{Dataset and model versions}

The evaluation uses a fixed-split decision-routing corpus constructed from named public NLI and stance-style sources, primarily SNLI, MultiNLI, and SemEval-style rumor/stance material, together with targeted hard-case examples designed to stress semantic ambiguity. The held-out validation and test splits remain fixed across the parent and refined model evaluations so that changes can be attributed to the refinement stage rather than to split variation.

The reported model is the refined model in this fixed-split evaluation sequence. It starts from a parent decision model and adds one boundary-refinement epoch using a small hard-example subset mined from prior diagnostic errors. Results are reported for the fixed validation and test splits:
\begin{itemize}[leftmargin=1.5em]
  \item validation: n = 44,404;
  \item test: n = 44,597.
\end{itemize}

The evaluation materials include full-split metrics, confusion matrices, calibration data, threshold sweeps, and reproducibility commands.

The primary metric is Macro F1 because the decision labels must be evaluated across all three outcomes rather than dominated by the largest class. Accuracy is reported as a secondary metric. Per-class precision, recall, and F1 are reported to expose tradeoffs among action, rejection, and deferral.

\subsection{Evaluation Questions}

The evaluation is organized around four questions:
\begin{enumerate}[leftmargin=1.5em]
  \item Does the model maintain balanced decision performance across YES, NO, and TBD?
  \item Does the deferral class behave as a learned outcome rather than as an afterthought to binary classification?
  \item Are the confidence scores sufficiently calibrated to support threshold review?
  \item Can an external reviewer identify the evaluated model version, data split, metrics, and commands used to reproduce the reported evidence?
\end{enumerate}

These questions are deliberately practical. A model used as a decision-control layer must be evaluated not only by aggregate score but also by the quality of its failure modes and the traceability of its evaluation evidence.

\subsection{Metrics and Their Roles}

Macro F1 measures whether performance is balanced across the bounded decision classes. Per-class F1 identifies which operational boundary is weakest. Precision and recall show whether each class is over-selected or under-selected. The confusion matrix exposes where errors move: for example, whether true TBD cases are being turned into action labels. Calibration error measures whether predicted confidence can be used in threshold policies. Threshold sweeps expose the coverage and deferral tradeoffs available to deployment owners.

The paper reports these metrics together because no single number captures the quality of a deferral-aware system. A higher aggregate F1 would be less useful if it concealed premature action on ambiguous inputs. Conversely, a conservative threshold could increase deferral while reducing automation. The evaluation evidence is intended to make these tradeoffs explicit.

\section{Results}

\subsection{Decision Performance}

\begin{center}
\begin{tabular}{lcc}
\toprule
Split & Accuracy & Macro F1 \\
\midrule
Validation (n=44,404) & 0.8224 & 0.8213 \\
Test (n=44,597) & 0.8260 & 0.8252 \\
\bottomrule
\end{tabular}
\end{center}

The validation and test results are close, which supports the stability of the reported behavior under the fixed data split. The test Macro F1 of 0.8252 should be interpreted together with the per-class table below because the operational meaning of errors differs across YES, NO, and TBD.

\subsection{Per-Class Test Performance}

\begin{center}
\begin{tabular}{lrrrr}
\toprule
Class & Precision & Recall & F1 & Support \\
\midrule
YES & 0.8205 & 0.8425 & 0.8314 & 14,883 \\
NO & 0.8598 & 0.8376 & 0.8486 & 15,650 \\
TBD & 0.7955 & 0.7958 & 0.7956 & 14,064 \\
\bottomrule
\end{tabular}
\end{center}

NO has the highest class F1 in the reported test result (F1 = 0.8486), indicating that the model learned a useful rejection boundary. TBD remains the hardest class (F1 = 0.7956), which is expected in a defer-aware system because the deferral region is defined by ambiguity rather than by a single semantic pattern. For this reason, the evaluation includes threshold sweeps and calibration evidence rather than relying only on a single argmax policy.

\subsection{Abstention and Coverage Baselines}

The full-coverage comparison separates learned deferral from forced binary classification. When the same outputs are collapsed into a binary YES/NO decision with no abstention, Macro F1 drops to 0.4945 and TBD F1 is zero because the model can no longer represent deferral. The learned three-way model at full coverage achieves Macro F1 = 0.8252.

Reject-threshold baselines answer a different question: what happens if a system abstains on a fraction of cases and evaluates only the retained subset? At 90\% retained coverage, confidence, entropy, and margin rejection produce similar Macro F1 values near 0.863. These numbers are not directly comparable to the full-coverage learned-TBD score because 10\% of cases are withheld from the retained set. They do show that thresholding remains a useful operating-policy layer on top of the learned distribution.

\begin{center}
\begin{tabular}{lrrrr}
\toprule
Method & Coverage & Macro F1 & YES F1 & TBD F1 \\
\midrule
Learned TBD, argmax & 1.00 & 0.8252 & 0.8314 & 0.7956 \\
Binary YES/NO, no abstention & 1.00 & 0.4945 & 0.7482 & 0.0000 \\
Confidence reject & 0.90 & 0.8631 & 0.8667 & 0.8358 \\
Entropy reject & 0.90 & 0.8624 & 0.8681 & 0.8329 \\
Margin reject & 0.90 & 0.8625 & 0.8662 & 0.8358 \\
\bottomrule
\end{tabular}
\end{center}

\begin{figure}[htbp]
\centering
\includegraphics[width=0.78\linewidth]{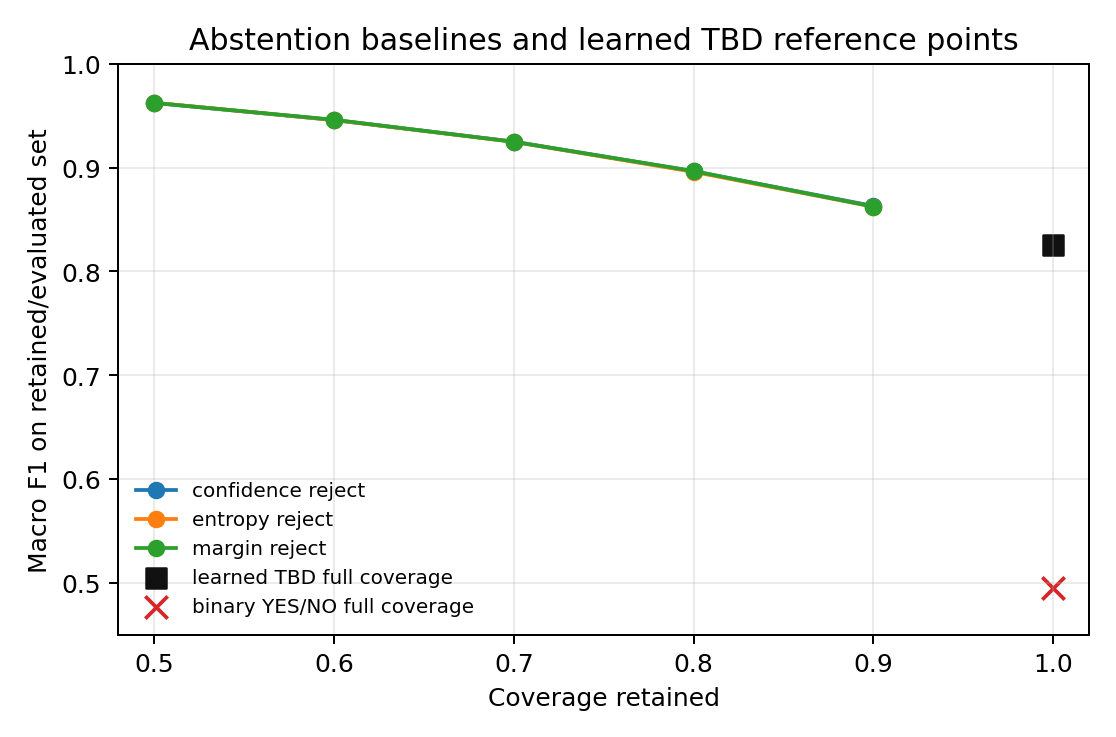}
\caption{Coverage-sensitive abstention baselines. Reject methods improve retained-set Macro F1 as coverage decreases; learned TBD and forced binary classification are shown as full-coverage reference points.}
\end{figure}

\subsection{Validation Confusion Matrix}

\begin{center}
\begin{tabular}{lccc}
\toprule
 & Pred YES & Pred NO & Pred TBD \\
\midrule
True YES & 12,496 & 999 & 1,467 \\
True NO & 1,035 & 13,066 & 1,433 \\
True TBD & 1,793 & 1,160 & 10,955 \\
\bottomrule
\end{tabular}
\end{center}

The validation confusion matrix shows that the largest residual issue is confusion between true TBD and direct action labels, especially TBD predicted as YES. This is the main operational boundary for future improvement: reducing premature action while preserving useful YES/NO coverage.

\subsection{Calibration and Stability}

The reported evaluation records validation ECE = 0.0338. This value indicates that the confidence distribution is suitable for inspection, but it should not be treated as a universal guarantee. Calibration is evaluated on the fixed validation split and should be rechecked after domain transfer or threshold changes.

The evaluation also includes a multi-seed stability check over validation seeds 42, 0, and 7, which produced identical Macro F1 values in the published validation procedure (std = 0.0). This is not a claim that all future training runs will be identical. It shows that the evaluated inference and scoring procedure is deterministic under the recorded configuration.

\subsection{Parent-to-Refined Model Effect}

The refined model should not be interpreted as a material raw-F1 improvement over its parent model. On the fixed test split, Macro F1 is essentially unchanged: parent model = 0.8254 and refined model = 0.8252. The measurable effect of the 323-row boundary-refinement subset is instead calibration and high-confidence error reduction. In the paired-comparison script, validation ECE improves from 0.0461 to 0.0413, test ECE improves from 0.0426 to 0.0411, and high-confidence error rate at threshold 0.85 drops from 0.0558 to 0.0485 on the test split. These paired-comparison ECE values are reported separately from the stored calibration report, which records validation ECE = 0.0338 under its stored binning procedure.

\begin{center}
\begin{tabular}{lrrrr}
\toprule
Metric & Parent val & Refined val & Parent test & Refined test \\
\midrule
Macro F1 & 0.8211 & 0.8213 & 0.8254 & 0.8252 \\
ECE & 0.0461 & 0.0413 & 0.0426 & 0.0411 \\
High-conf. error @0.85 & 0.0574 & 0.0499 & 0.0558 & 0.0485 \\
TBD rate & 0.3183 & 0.3120 & 0.3221 & 0.3155 \\
\bottomrule
\end{tabular}
\end{center}

\begin{figure}[htbp]
\centering
\includegraphics[width=0.78\linewidth]{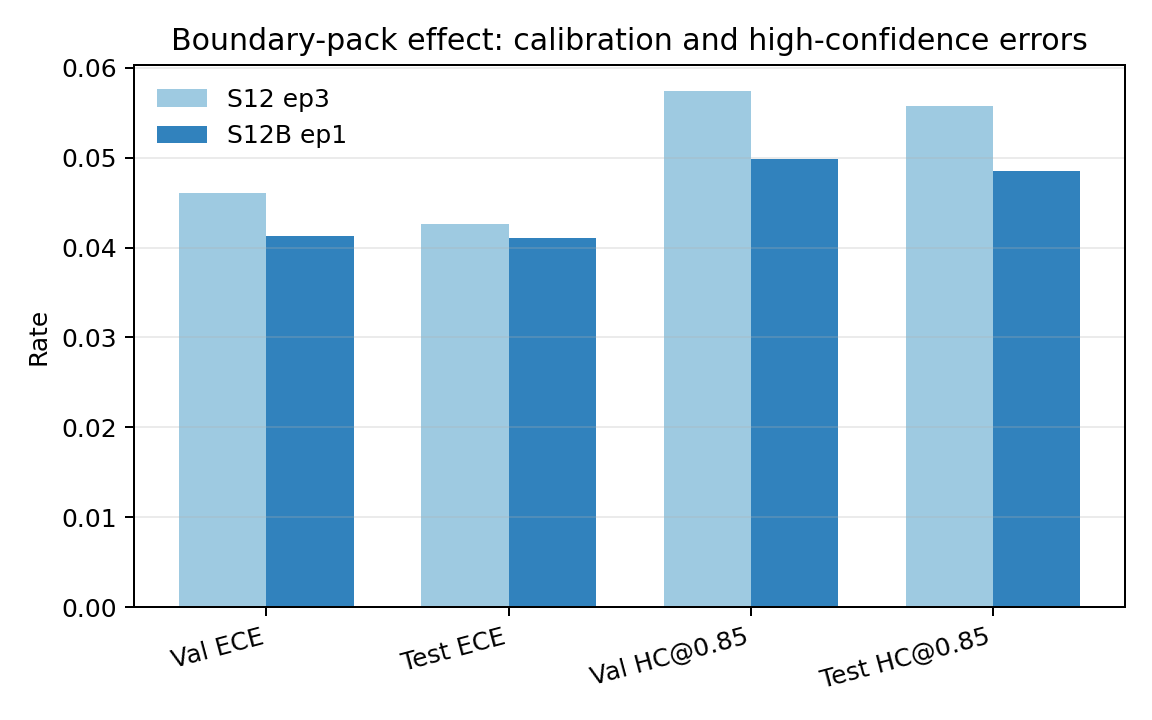}
\caption{Parent-to-refined comparison. The boundary-refinement pass leaves Macro F1 nearly flat but reduces calibration error and high-confidence error rates.}
\end{figure}

\subsection{Interpretation of the Remaining Error Mode}

The most important residual error is not a simple lack of class balance. YES, NO, and TBD have comparable support in the test split, and the model performs strongly on NO. The remaining difficulty is semantic: a true TBD case may resemble a YES or NO case because it shares terms, roles, or surface structure while lacking decisive evidence. This is why the boundary-refinement subset emphasizes contradiction, negation, role and number swaps, high-overlap non-entailment, and hedged or modal statements.

From an operational perspective, the highest-risk error is a true TBD routed as YES or NO. Such cases reduce the value of deferral because they turn ambiguous inputs into direct actions. The threshold-sweep outputs are therefore part of the evaluation evidence used to select an operating point.

The qualitative audit on the full test split records 7,758 errors out of 44,597 examples (17.4\%). The largest error categories are TBD routed to YES (1,729), NO routed to TBD (1,528), YES routed to TBD (1,350), and TBD routed to NO (1,143). Hard binary confusions are smaller but operationally important: YES routed to NO occurs 994 times and NO routed to YES occurs 1,014 times. The audit examples indicate that the residual errors concentrate around lexical overlap, missing evidence, negation/polarity ambiguity, and cases where a plausible inference goes beyond the premise.

Examples from the full test-set audit illustrate the boundary. In a TBD$\rightarrow$YES case, ``well maybe something will open up for you'' was paired with ``maybe there will be an opening for you in the company''; the model predicted YES with probability 0.564, while the true label was TBD because the statement remains modal. In a TBD$\rightarrow$NO case, ``clever man, Bauerstein'' was paired with ``Bauerstein was an old man''; the model predicted NO with probability 0.715, while the premise does not establish age. In a NO$\rightarrow$TBD case, ``Whillan Beck'' referred to a water source powering a mill, while the hypothesis treated it as the man who built the mill; the model deferred despite a contradiction. These examples show why the residual boundary is semantic rather than purely lexical.

\begin{center}
\begin{tabular}{lrrr}
\toprule
Error category & Count & Mean confidence & Mean margin \\
\midrule
TBD$\rightarrow$YES & 1,729 & 0.7391 & 0.5386 \\
TBD$\rightarrow$NO & 1,143 & 0.6698 & 0.4125 \\
YES$\rightarrow$TBD & 1,350 & 0.7207 & 0.5063 \\
NO$\rightarrow$TBD & 1,528 & 0.7376 & 0.5355 \\
YES$\rightarrow$NO & 994 & 0.7059 & 0.4887 \\
NO$\rightarrow$YES & 1,014 & 0.7165 & 0.5107 \\
\bottomrule
\end{tabular}
\end{center}

\begin{figure}[htbp]
\centering
\begin{minipage}{0.48\linewidth}
\centering
\includegraphics[width=\linewidth]{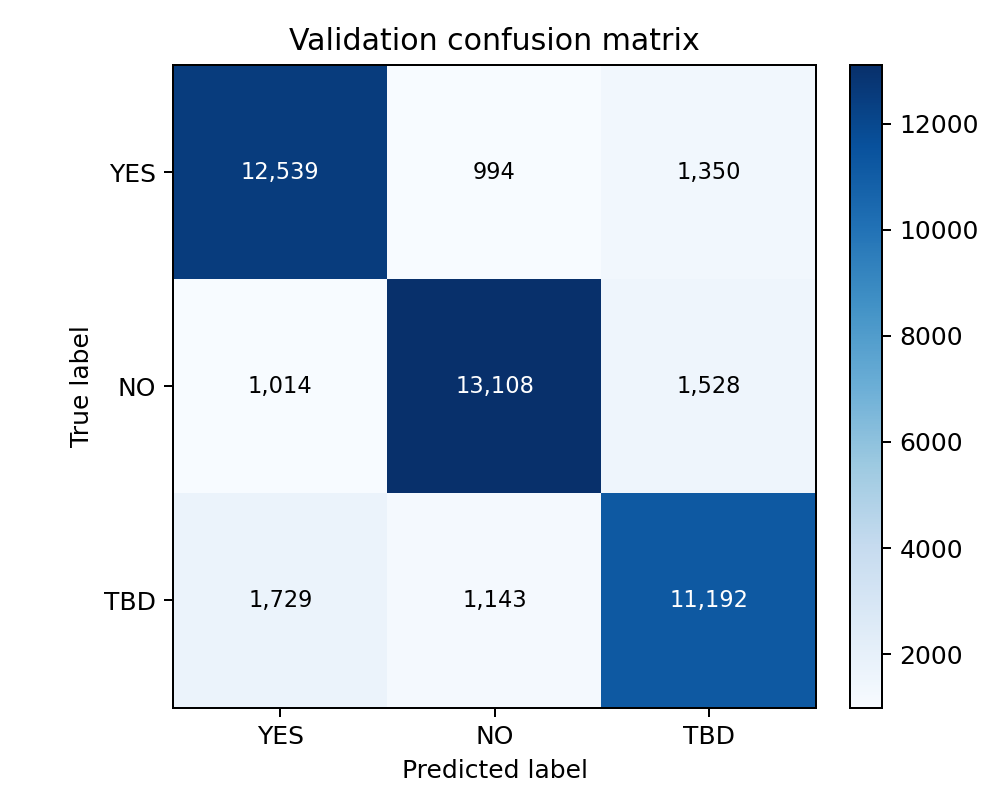}
\end{minipage}
\hfill
\begin{minipage}{0.48\linewidth}
\centering
\includegraphics[width=\linewidth]{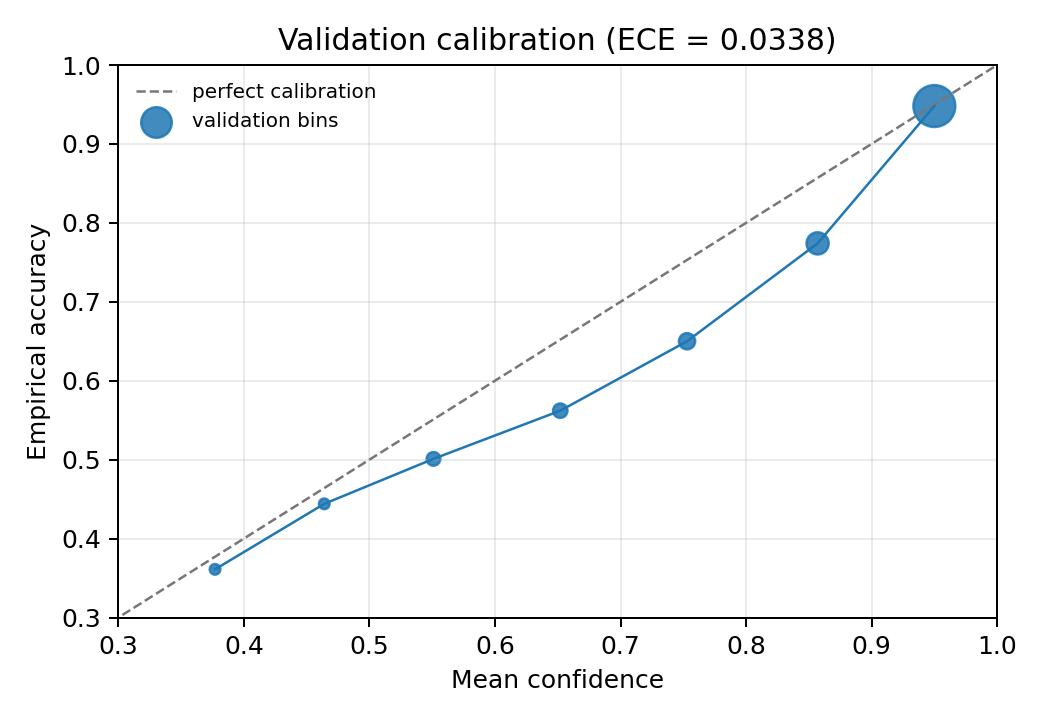}
\end{minipage}
\caption{Evaluation evidence for the reported model version. Left: validation confusion matrix. Right: validation calibration curve with ECE = 0.0338.}
\end{figure}

\section{Audit-Oriented Evaluation Evidence}

The model version is documented with audit-oriented evidence in addition to evaluation scores. For the reported model version, the public release includes an evidence directory containing the following materials:
\begin{itemize}[leftmargin=1.5em]
  \item evaluation and calibration reports;
  \item calibration data with validation ECE = 0.0338;
  \item threshold-sweep outputs for TBD-fallback operating policies;
  \item multi-seed stability evidence, with validation seeds 42, 0, and 7 producing identical Macro F1 (std = 0.0);
  \item abstention, dataset, annotation, deferral-error, and parent/refined comparison reports;
  \item run metrics, confusion matrix, diagnostics, and reproducibility commands.
\end{itemize}

This structure is part of the evaluation contribution. It exposes the evidence needed to inspect operating behavior rather than only the final model version.

\begin{center}
\begin{tabular}{ll}
\toprule
Evidence type & Review purpose \\
\midrule
Metrics reports & Verify full-split accuracy, Macro F1, and per-class behavior \\
Confusion matrices & Inspect movement among YES, NO, and TBD outcomes \\
Calibration data & Assess whether confidence supports threshold policies \\
Threshold sweeps & Select operating points for automation versus deferral \\
Stability checks & Confirm deterministic evaluation behavior for the release \\
Reproducibility commands & Reconstruct the evaluation procedure \\
Post-training tests & Review abstention, dataset, and error evidence \\
\bottomrule
\end{tabular}
\end{center}

\subsection{Why Audit Evidence Matters}

For decision-control models, reproducibility and auditability are not secondary documentation tasks. They determine whether a model can be reviewed before deployment. A headline score does not show whether the confidence distribution is calibrated, whether thresholds were selected after inspecting tradeoffs, whether results are stable across seeds, or whether the exact evaluated model version can be located.

The evaluation is therefore recorded as a set of review evidence. The evidence directory is designed to answer reviewer questions: Which model version was evaluated? Which data split was used? What were the per-class outcomes? What threshold sweeps were run? How stable are the results? Where are the reproducibility commands? This makes the reported result more transparent than a standalone benchmark table.

\section{Discussion}

The results show that bounded decision control can be evaluated as a decision interface rather than as ordinary classification alone. The reported model achieves balanced performance across YES, NO, and TBD, with NO F1 = 0.8486 and TBD F1 = 0.7956 on the full test split. The remaining difficulty is the boundary between actionable decisions and deferral, especially ambiguous cases where a human reviewer or downstream policy may prefer conservative routing.

Calibration, threshold sweeps, and multi-seed stability make the model's operating behavior inspectable. Downstream systems can select operating thresholds and document tradeoffs without retraining the core model. Operationally, decision control refers to separating the model's bounded distribution from the deployment policy that maps scores to routing decisions via recorded thresholds and fallback rules (and, in future validated model versions, auxiliary decision-relative semantic channels).

The auxiliary architecture also matters, but empirical claims should remain scoped. In the reported evaluation, decision behavior is validated; emotion metrics are not reported because the emotion head is disabled. Future releases can separately validate auxiliary value and emotion behavior as independent outputs.

The result is not an isolated leaderboard metric. It is a learned-deferral decision model evaluated with evidence that describes classwise behavior, calibration, threshold tradeoffs, stability, and model identity. These materials are necessary when a model is used as a decision interface rather than a passive classifier.

\subsection{Interpretation of the Score Plateau}

The reported result is close to the preceding parent model in raw Macro F1, but it is better supported by evaluation evidence because the decision metrics are accompanied by corrected evaluation behavior, calibration evidence, stability checks, abstention baselines, error audits, and a clearer evidence structure. This distinction matters. A slightly higher score without reliable evaluation handling, traceable evaluation evidence, or threshold evidence would be less informative for external review.

The observed plateau near the low-0.82 Macro F1 range suggests that remaining gains are concentrated in the semantic deferral boundary rather than in generic optimization. The model already distinguishes NO well. The hard cases are those where the input contains familiar words or plausible structure but does not supply enough evidence for action. Further improvement is therefore more likely to come from better boundary labeling, domain-specific validation, longer-context handling, and operating-threshold selection than from simply increasing dataset size.

\subsection{Implications for Decision Workflows}

The practical implication is that AI decision systems should expose uncertainty as a routable outcome. A deployment owner can then decide how uncertain cases move through the workflow: manual review, evidence request, policy escalation, or rejection by default. This is different from treating uncertainty as a hidden property of a model confidence score. It also makes operational control explicit: changing the workflow response to uncertainty becomes a documented policy change, not a hidden model change.

The model makes this separation explicit. It learns a deferral class, while the operating policy selects thresholds and routing behavior. This separation supports review because the model and the deployment rule can be examined independently. If an organization changes its risk tolerance, it can adjust thresholds and document the change without altering the underlying model.

\section{Limitations}

Results are specific to the evaluated corpus, the reported model version, and the evaluation setup described here. The control interface is domain-agnostic, but the reported performance is not a domain-transfer result. Additional domains require separate validation, calibration review, threshold selection, policy-version review, and error audit. Inputs are truncated to 128 tokens, so long-context use cases require chunking or preprocessing. The emotion head is disabled in this evaluation, so this paper does not claim validated emotion-head performance for the reported model version. The dataset composition includes public NLP sources and curated decision examples, but raw training data is not redistributed in this repository. Users should consult original dataset licenses before reusing source datasets.

The paper also does not claim that TBD is always the correct action under uncertainty. TBD is a model output and an operating-policy option, not a substitute for domain governance. Deployment owners must decide how deferred cases are routed and what thresholds are acceptable for their risk environment. Similarly, the reported calibration value is tied to the validation split and should be remeasured after domain transfer, threshold changes, or additional fine-tuning.

The abstention experiments are diagnostic baselines, not a complete comparison against independently trained abstention systems. The forced binary baseline tests the loss of the TBD class, while confidence, entropy, and margin rejection test retained-set behavior under thresholded abstention. A more complete follow-up study should include separately trained binary and selective-classification baselines, conformal abstention, and domain-transfer evaluations.

Finally, the audit evidence should be understood as a review aid, not as a guarantee of universal safety. It makes the evidence inspectable, but each deployment still requires application-specific validation and monitoring.

\section{Future Work}

Three extensions would strengthen the research claim beyond the present study. First, independently trained baselines should be added, including binary classifiers with rejection, selective-classification objectives, conformal prediction wrappers, and abstention-trained models. Second, domain-transfer evaluations should test whether learned deferral and calibration behavior hold outside the evaluated corpus, for example in policy approval, moderation escalation, compliance review, financial triage, or enterprise workflow routing. Third, auxiliary value and emotion heads should be re-enabled and evaluated as independent decision-relative semantic outputs before they are treated as validated model claims. A further deployment study should evaluate whether auxiliary channels can be used as real-time policy controls while preserving auditability, calibration, and separation between model scores and operating policy.

\section{Conclusion}

This paper presented a bounded decision-control model with learned deferral. The motivating problem is that operational AI systems often need to decide whether to act, reject, or defer, and a forced binary or generative output can hide uncertainty from the surrounding workflow. The model addresses this problem with a three-way learned decision interface: YES, NO, and TBD.

On the full fixed test split, the reported model achieves Accuracy = 0.8260 and Macro F1 = 0.8252, with per-class F1 scores of 0.8314 (YES), 0.8486 (NO), and 0.7956 (TBD). A forced binary YES/NO view of the same task drops to Macro F1 = 0.4945 because it cannot represent deferral. The boundary-refinement pass does not materially improve raw F1 over its parent model, but it reduces calibration error and high-confidence error rates. The result is accompanied by calibration evidence, threshold sweeps, stability checks, confusion matrices, abstention baselines, error audits, and reproducibility materials. The paper's claim is therefore not only that the model reaches a particular score, but that learned deferral can be evaluated as an auditable decision-control interface whose inference-time behavior can be governed through documented operating policy.

The remaining work is clear: improve the semantic boundary between direct action and deferral, validate auxiliary value and emotion outputs as independent empirical claims, test auxiliary channels as policy controls, and repeat calibration and threshold review for each deployment domain. Within the evaluated scope, the reported model provides a concrete model version for bounded decision routing under ambiguity.

\bibliographystyle{plain}
\bibliography{references}

\begin{thebibliography}{10}

\bibitem{barbieri2020tweeteval}
Francesco Barbieri, Jose Camacho-Collados, Luis Espinosa~Anke, and Leonardo
  Neves.
\newblock Tweeteval: Unified benchmark and comparative evaluation for tweet
  classification.
\newblock In {\em Findings of the Association for Computational Linguistics:
  EMNLP 2020}, pages 1644--1650, 2020.

\bibitem{bowman2015snli}
Samuel~R. Bowman, Gabor Angeli, Christopher Potts, and Christopher~D. Manning.
\newblock A large annotated corpus for learning natural language inference.
\newblock In {\em Proceedings of the 2015 Conference on Empirical Methods in
  Natural Language Processing}, pages 632--642, 2015.

\bibitem{caruana1997multitask}
Rich Caruana.
\newblock Multitask learning.
\newblock {\em Machine Learning}, 28:41--75, 1997.

\bibitem{demszky2020goemotions}
Dorottya Demszky, Dana Movshovitz-Attias, Jeongwoo Ko, Alan Cowen, Gaurav
  Nemade, and Sujith Ravi.
\newblock Goemotions: A dataset of fine-grained emotions.
\newblock In {\em Proceedings of the 58th Annual Meeting of the Association for
  Computational Linguistics}, pages 4040--4054, 2020.

\bibitem{devlin2019bert}
Jacob Devlin, Ming-Wei Chang, Kenton Lee, and Kristina Toutanova.
\newblock Bert: Pre-training of deep bidirectional transformers for language
  understanding.
\newblock In {\em Proceedings of NAACL-HLT}, 2019.

\bibitem{elyaniv2010selective}
Ran El-Yaniv.
\newblock On the foundations of noise-free selective classification.
\newblock {\em Journal of Machine Learning Research}, 11:1605--1641, 2010.

\bibitem{franc2023reject}
Vojtech Franc, Daniel Prusa, and Vojtech Voracek.
\newblock Optimal strategies for reject option classifiers.
\newblock {\em Journal of Machine Learning Research}, 24, 2023.

\bibitem{geifman2017selective}
Yonatan Geifman and Ran El-Yaniv.
\newblock Selective classification for deep neural networks.
\newblock In {\em Advances in Neural Information Processing Systems}, 2017.

\bibitem{go2009twitter}
Alec Go, Richa Bhayani, and Lei Huang.
\newblock Twitter sentiment classification using distant supervision.
\newblock Technical report, Stanford University, 2009.

\bibitem{guo2017calibration}
Chuan Guo, Geoff Pleiss, Yu~Sun, and Kilian~Q. Weinberger.
\newblock On calibration of modern neural networks.
\newblock {\em Proceedings of the 34th International Conference on Machine
  Learning}, 2017.

\bibitem{hendrycks2021ethics}
Dan Hendrycks, Collin Burns, Steven Basart, Andrew Critch, Jerry Li, Dawn Song,
  and Jacob Steinhardt.
\newblock Aligning ai with shared human values.
\newblock In {\em International Conference on Learning Representations}, 2021.

\bibitem{hoover2020mftc}
Joe Hoover, Gwenyth Portillo-Wightman, Leigh Yeh, Shreya Havaldar,
  Aida~Mostafazadeh Davani, Ying Lin, Brendan Kennedy, Mohammad Atari, Zahra
  Kamel, Madelyn Mendlen, Gabriela Moreno, Christina Park, Tingyee~E. Chang,
  Jenna Chin, Christian Leong, Jun~Yen Leung, Arineh Mirinjian, and Morteza
  Dehghani.
\newblock Moral foundations twitter corpus: A collection of 35k tweets
  annotated for moral sentiment.
\newblock {\em Social Psychological and Personality Science}, 11(8):1057--1071,
  2020.

\bibitem{maas2011learning}
Andrew~L. Maas, Raymond~E. Daly, Peter~T. Pham, Dan Huang, Andrew~Y. Ng, and
  Christopher Potts.
\newblock Learning word vectors for sentiment analysis.
\newblock In {\em Proceedings of the 49th Annual Meeting of the Association for
  Computational Linguistics: Human Language Technologies}, pages 142--150,
  2011.

\bibitem{malo2014good}
Pekka Malo, Ankur Sinha, Pekka Korhonen, Jyrki Wallenius, and Pyry Takala.
\newblock Good debt or bad debt: Detecting semantic orientations in economic
  texts.
\newblock {\em Journal of the Association for Information Science and
  Technology}, 65(4):782--796, 2014.

\bibitem{poria2019meld}
Soujanya Poria, Devamanyu Hazarika, Navonil Majumder, Gautam Naik, Erik
  Cambria, and Rada Mihalcea.
\newblock Meld: A multimodal multi-party dataset for emotion recognition in
  conversations.
\newblock In {\em Proceedings of the 57th Annual Meeting of the Association for
  Computational Linguistics}, pages 527--536, 2019.

\bibitem{rashkin2019empathetic}
Hannah Rashkin, Eric~Michael Smith, Margaret Li, and Y-Lan Boureau.
\newblock Towards empathetic open-domain conversation models: A new benchmark
  and dataset.
\newblock In {\em Proceedings of the 57th Annual Meeting of the Association for
  Computational Linguistics}, pages 5370--5381, 2019.

\bibitem{rosenthal2017semeval}
Sara Rosenthal, Noura Farra, and Preslav Nakov.
\newblock Semeval-2017 task 4: Sentiment analysis in twitter.
\newblock In {\em Proceedings of the 11th International Workshop on Semantic
  Evaluation (SemEval-2017)}, pages 502--518, 2017.

\bibitem{sap2020socialbias}
Maarten Sap, Saadia Gabriel, Lianhui Qin, Dan Jurafsky, Noah~A. Smith, and
  Yejin Choi.
\newblock Social bias frames: Reasoning about social and power implications of
  language.
\newblock In {\em Proceedings of the 58th Annual Meeting of the Association for
  Computational Linguistics}, pages 5477--5490, 2020.

\bibitem{schwartz2012overview}
Shalom~H. Schwartz.
\newblock An overview of the schwartz theory of basic values.
\newblock {\em Online Readings in Psychology and Culture}, 2(1), 2012.

\bibitem{williams2018broad}
Adina Williams, Nikita Nangia, and Samuel~R. Bowman.
\newblock A broad-coverage challenge corpus for sentence understanding through
  inference.
\newblock In {\em Proceedings of the 2018 Conference of the North American
  Chapter of the Association for Computational Linguistics: Human Language
  Technologies, Volume 1 (Long Papers)}, pages 1112--1122, 2018.

\bibitem{ziems2023normbank}
Caleb Ziems, Jane Dwivedi-Yu, Yi-Chia Wang, Alon Halevy, and Diyi Yang.
\newblock Normbank: A knowledge bank of situational social norms.
\newblock In {\em Proceedings of the 61st Annual Meeting of the Association for
  Computational Linguistics (Volume 1: Long Papers)}, pages 7756--7776, 2023.

\end{thebibliography}

\end{document}